# A Study on the Refining Handwritten Font by Mixing Font Styles

Avinash Kumar, Kyeolhee Kang, Ammar ul Hassan, Jaeyoung Choi
School of Computer Science & Engineering
Soongsil University
Seoul, South Korea
{kumaravinashsw44, kyeolhee.ss, ammar.instantsoft}@gmail.com, choi@ssu.ac.kr

*Abstract*— **Handwritten fonts have a distinct expressive character, but they are often difficult to read due to unclear or inconsistent handwriting. FontFusionGAN (FFGAN) is a novel method for improving handwritten fonts by combining them with printed fonts. Our method implements generative adversarial network (GAN) to generate font that mix the desirable features of handwritten and printed fonts. By training the GAN on a dataset of handwritten and printed fonts, it can generate legible and visually appealing font images. We apply our method to a dataset of handwritten fonts and demonstrate that it significantly enhances the readability of the original fonts while preserving their unique aesthetic. Our method has the potential to improve the readability of handwritten fonts, which would be helpful for a variety of applications including document creation, letter writing, and assisting individuals with reading and writing difficulties. In addition to addressing the difficulties of font creation for languages with complex character sets, our method is applicable to other text-image-related tasks, such as font attribute control and multilingual font style transfer.**

## I. Introduction

Handwritten fonts are a unique and expressive form of communication. They can be used to add a personal touch to documents, letters, and other creative projects. However, handwritten fonts can also be difficult to read, especially if the handwriting is not clear or consistent.

Font creation poses a significant challenge, particularly for languages characterized by a large number of complex characters. For instance, the Korean language encompasses 11,172 Hangul characters. While the Chinese dictionary comprises over 50,000 Hanja characters. Recent advancements in generative models have paved the way for novel font synthesis methods [1,3,11,12], leveraging generative adversarial networks (GANs). These methods approach font synthesis problem as an image-to-image translation task, training in either a supervised setting, involving paired image data [1,3,11,12], or a set level, incorporating font style labels.

Despite the impressive achievements of existing methods in generating realistic fonts, they exhibit certain limitations. Firstly, these methods heavily rely on large sets of paired training samples. Secondly, they often necessitate additional fine-tuning steps to effectively learn previously unseen font styles. Thirdly, their applicability is generally restricted to specific language characters. Lastly, most of the existing methods are doing style transfer (transfer the style of reference image into the content image) not the actual style mixing. In style transfer we lose the actual unique quality and style of content image.

To overcome these limitations, we propose a straightforward yet highly effective approach for style mixing and enhancing handwritten font images. Our approach draws inspiration from the mixing regularization technique employed in the state-of-the-art method for generating high-resolution face images, namely StyleGAN [8]. By utilizing mixing regularization, we facilitate style localization by generating images using two random latent codes, instead of employing only one, during the training process.

In this paper, we propose a FontFusionGAN (FFGAN) for refining handwritten fonts by mixing them with printed fonts. Our method uses a generative adversarial network (GAN) to generate new fonts that combine the best features of both handwritten and printed fonts. The FFGAN is trained on a dataset of handwritten and printed fonts, and it learns to generate new fonts that are both visually appealing and legible. We evaluated our method on a dataset of handwritten fonts, the results showed that our method was able to generate new fonts images that were significantly better than the original handwritten fonts with their own unique style. Our model has the potential to be used to improve the readability of handwritten fonts through font fusion. It could also be used to create new fonts for documents, letters, creative projects, and provide help to people with dyslexia or other reading difficulties.

Our experiments show that our approach can generate high-quality mixed font images with a variety of different styles. We also show that our approach can be used to refine existing fonts, making them more visually appealing. We believe that our approach is a promising new method for making font fusion and improves the quality of handwritten font images. It has potential to make font creation more efficient and easier to use, and it can be used to create a wider variety of mixed fonts.

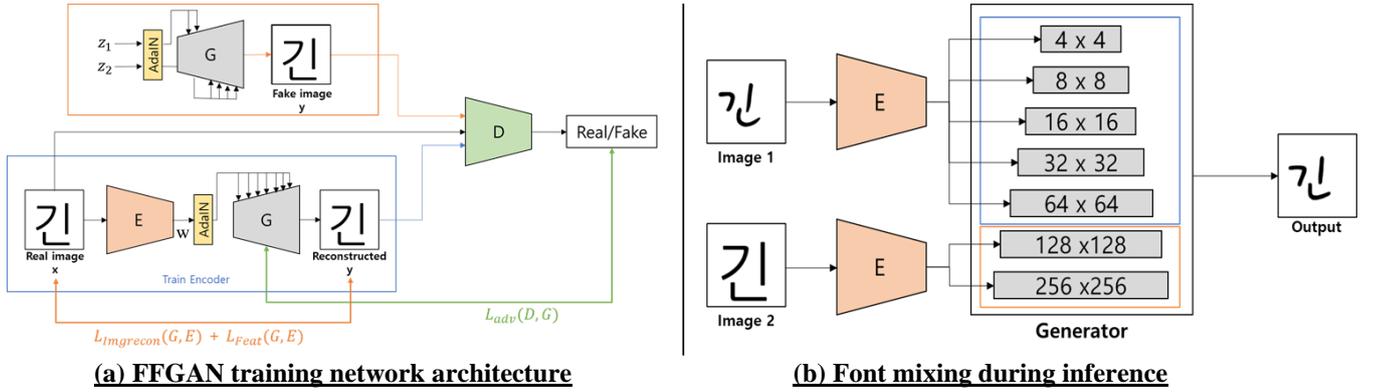

**(a) FFGAN training network architecture**     **(b) Font mixing during inference**

Fig. 1. Illustrates the main architecture of our proposed FFGAN. In (a), the training scheme is shown, where FFGAN learns two key properties: style mixing (upper orange box) and projecting input images to style latent space for accurate reconstruction (lower blue box). During style mixing learning, random noise vectors ($z^1$, $z^2$) are used in Generator G to generate fake image y. A fusion encoder E is trained to map real images x to corresponding style latent space (W), which is then fed into G for image reconstruction y. In (b), the inference stage demonstrates controllable font mixing using the trained E and G. Content image (image 1) and style image (image 2) are processed by E to extract respective style latent vectors ($W^1$, $W^2$). These vectors are then fed into the low and high-resolution layers of G, generating an output image with content from (Image 1) and style from (Image 2).

We introduced FontFusionGAN (FFGAN), which enhances the StyleGAN model by incorporating a fusion encoder network and by removing the mapping network. This addition enables FFGAN to learn the process of projecting any given font image into its corresponding style latent space. Our fusion encoder accurately performs this fusion in real time, generating a latent space representation compatible with the StyleGAN generator. As a result, FFGAN enables controllable style mixing for refining handwritten font images during inference, using inject index property we can input the handwritten font image into the initial 6 layers of generator and printed font image into the last layer of generator which shows the better-quality results of fusion.

In contrast to image-to-image (I2I) based font synthesis methods that employ separate encoders and multitask discriminators for style and content learning and for that, so these methods learn complete style transfer on the content images, FFGAN jointly learns that style and content are independent of each other. This simple yet effective strategy allows FFGAN to fuse previously unseen font styles in a few-shot setting without any architectural constraints or additional training requirements. Moreover, FFGAN operates in an unsupervised manner, eliminating the need for image or class label supervision such as character content or font style labels.

We conducted various experiments to validate our proposed method quantitatively and qualitatively. Furthermore, by combining our unsupervised approach with a novel training setup, FFGAN can be easily extended to other text-image-related task, including style transfer, multilingual font style transfer, font attribute control, and random font style generation. These extensions demonstrate the generalization and potential of our proposed method.

## II. FFGAN

### A. Architecture overview

The overall architecture of FFGAN is illustrated in Fig. 1(a). The framework comprises three main components: a generator G, a fusion encoder E, and a discriminator D.

Generator G begins with a constant vector of size $4 \times 4 \times 64$, which represents the spatial dimensions of the initial layers. In our framework, the style vector for the generator G can be obtained from two different sources. Firstly, from the orange box in Fig. 1(a), two latent codes $z^1$ and $z^2$ are sampled from a Gaussian distribution. Secondly, from the blue box in Fig. 1(a), a real image x is passed through the fusion encoder E, which encodes it to produce a style vector W. These latent vectors ($z^1$, $z^2$) or a style vector (W) in then transformed and injected into each block of the generator G after the convolution layers using adaptive instance normalization (AdaIN) [7]. Unlike existing font generation methods that employ a multi-task discriminator for generating realistic images and content-style supervision, we use a discriminator D for adversarial training. Discriminator D role is to determine whether an input image is real or fake.

### B. Style mxing at inference

Using the trained FFGAN model, we can generate a mixed font set that combines both handwritten and printed through mixing regularization. To achieve this, we extract style vectors $W^1$ and $W^2$ from a Handwritten image (Image 1) and a Printed image (Image 2), respectively.

Next, we inject $W^1$ in the first 6 layers and $W^2$ into the last layer of the generator. This injection of style vectors at different layers ensures that the generated mixed font retains the desired content structure from the content image while incorporating style attributes, such as strokes and thickness, from the style image.

*C. Objective functions*

We used the following objective functions, explained next, to enable the learning of controllable style mixing for font mixing regularization.

*1) Adversarial loss*

In the adversarial training process of the generator G and discriminator D, we utilized a non-saturating GAN loss approach. The discriminator's role is to classify the generated fake image (y) produced by the generator, which can be randomly generated from a Gaussian distribution ($z^1$, $z^2$) or reconstructed from the input image (x) as a fake image after passing through a transformation (x -> W). On the other hand, the generator aims to deceive the discriminator by synthesizing images that appear realistic.

$$\min_D \max_G L_{adv}(D, G) = E_x[\log D(x)] + E_y[-\log(D(y))] \quad (1)$$

*2) Image reconstruction loss*

This loss function was utilized to facilitate image reconstruction learning. We minimized the pixel-level disparity between the actual image x and the corresponding reconstructed image y using the $L_1$ distance measurement. This loss was applied to both the generator G and the encoder E during joint training. The specific formulation of the loss can be represented as follows:

$$L_{Imgrecon}(G, E) = E_{x,y}[\|x-y\|_1] \quad (2)$$

*3) Feature matching loss*

In our research, we noticed that using pixel-level loss alone often leads to the production of blurry images. To overcome this issue, we incorporated a feature matching loss into our approach. To this this, we utilized the discriminator D as a feature extractor, which we referred to as $D^f$. By obtaining the intermediate feature maps for both the input image x and the generated image y using $D^f$, we could compare then using the $L_1$ distance. This comparison allowed us to minimize the discrepancies between the two images. We applied this loss to both the generator G and the encoder E. Thus, our enhanced loss formulation helps in generating smoother and more visually appealing images.

$$L_{Feat}(G, E) = E_x[\|D^f(x) - D^f(y)\|_1] \quad (3)$$

The final learning objective of the proposed FFGAN is as follows:

$$\min_D \max_G \lambda_{adv} L_{adv}(D, G) + \lambda_{imgrecon} L_{imgrecon}(G, E) + \lambda_{feat} L_{feat}(G, E) \quad (4)$$

and $\lambda_{adv}$, $\lambda_{imgrecon}$ and $\lambda_{feat}$ are the tunable hyperparameters.

### III. EXPERIMENT

In this section, we validate our proposed FFGAN for Chinese and Korean font mixing tasks.

*A. Datasets*

For the Chinese Font mixing task evaluation, we collected a diverse dataset from freechinesefont. 110 font style 55 handwritten and 55 printed are included. Each font has the 1000 most common Chinese characters. Training was 80% and testing 20%. The test set has 200 unseen training font characters and 20% of the test front for generalization ability. The training set has 800 characters per font.

For the Korean hangul font mixing task, we created a dataset of 174 Google fonts with an equal distribution of handwritten and printed fonts. Each font has the most common 2350 Korean hangul characters. 80% was for training and 20% for testing. The training set had 2000 characters and 20% unseen testing fonts. The images were standardized to a size of 128 × 128 pixels.

*B. Training Details*

Our generator G and discriminator D were based on the StyleGAN model [9], but the following details were modified. The non-linear mapping network is removed from the network. The dimensions of the latent vectors $z^1$ and $z^2$ were set to 64. The channel sizes of generator G and discriminator D were reduced by half. We did not use any regularization in G. Our discriminator D also extracted features from the intermediate layers for training fusion encoder E and generator G. All the weights in the loss functions $\lambda_{adv}$, $\lambda_{imgrecon}$, $\lambda_{feat}$ (Eq. 4) was set to 1, that is $\lambda = 1$. For the rest, we followed the default setting of StyleGAN, such as $R_1$ regularization in D, Adam optimizer [10], learning rates and the exponential moving average of the generator. For fusion encoder E, we used the same architecture as D up to the 8 × 8 layers and added an average pooling layer, which proved to be stable and generalized. In contrast to discriminator D, fusion encoder E did not employ minibatch discrimination.

To train the entire system, we jointly trained generator G, the fusion encoder E, and the discriminator D. We assumed that there was no supervision at the image or set levels. To force the generator to learn style mixing, we sampled two latent vector $z^1$ and $z^2$ from the gaussian distribution then applied $z^1$ before and $z^2$ after a crossover point (orange box of Fig. 1 (a)). Injecting latent vector $z^1$ in low resolution layers controls the overall shape of the glyph, while injecting latent vector $z^2$ in high-resolution layers controls the glyph font style like thickness, strokes, and sheriffs. However, when the style vector W originates from fusion encoder E while learning to project the input image to the corresponding latent space for accurate reconstruction, it is injected directly into all layers of G (blue box of Fig. 1(a)).

*C. Qualitative results*

The visual outcomes are illustrated in Fig. 2, showcasing the qualitative results. We generate diverse Korean and Chinese characters by employing various font styles that have not been seen before. A key concept behind our FFGAN model is to achieve disentanglement (style mixing) between handwritten and printed fonts using mixing regularization in a completely unsupervised manner. During the inference stage, the fusion encoder can be utilized to extract style codes from the glyph images, enabling the performance of style mixing. We observe that our proposed approach effectively separates style and content, demonstrating its effectiveness on the mixed images. In Fig. 3, we tested our model with our own handwriting images which shows very good results refining it's also proofed the few shot learning behavior of our model.

Fig. 2. Font Fusion example. FFGAN can successfully generate a glyph image (output) with the handwritten font image and printed font image.

IV. CONCLUSION

In this paper, we propose a novel approach for refining handwritten fonts using mixing regularization technique. Our method eliminates the need for manual supervision, offering greater control over content and style in front synthesis. Through extensive experiments and evaluations, we demonstrated the superiority of our approach in terms of legibility, diversity, and style consistency. This research opens up new possibilities for efficient and high-quality font generation, with potential applications in various text-image-related problems.

Fig. 3. Represents the real handwriting images of human (from our team) we mix them with printed fonts and get a refined font.

*Acknowledgement*

This work was supported by the Soongsil University Research Fund of 2022.